\documentclass[journal]{IEEEtran}

\usepackage{cite}

\ifCLASSINFOpdf
\else
\fi

\usepackage{multirow}
\usepackage{pifont}
\usepackage{amsmath}
\usepackage{algorithm}
\usepackage{algorithmic}

\usepackage{array}

\usepackage{bm}

\usepackage{url}

\usepackage{graphicx}

\usepackage{subfigure}
\usepackage{booktabs}

\begin{document}
%
\title{One to Multiple Mapping Dual Learning: Learning Multiple Sources from One Mixing Signal}

\author{Ting Liu,
        Wenwu Wang,
        Xiaofei Zhang,
        Jingwen Yan,
        and Yina Guo
\thanks{Ting Liu,
        Xiaofei Zhang,
        Jingwen Yan,
        and Yina Guo are with the Department of Electronics and Communication Engineering, Taiyuan University of Science and Technology, Taiyuan, 030024, China. Corresponding author's e-mail: zulibest@163.com.

        Wenwu Wang is with the Centre for Vision, Speech and Signal Processing,
University of Surrey, Guildford, Surrey GU2 7XH, U.K. Ting Liu and Wenwu Wang contributed equally to this work. (e-mail: liuting328511@163.com, w.wang@surrey.ac.uk)

Jingwen Yan is with the Guangdong Provincial Key Laboratory of Digital Signal and Image Processing Technology.
}
}

\maketitle

\begin{abstract}
Single channel blind source separation (SCBSS) refers to separating multiple sources from a mixing signal collected by a single sensor. Existing methods for SCBSS mainly focus on separating two sources and have limited generalization performance. To address these problems, an algorithm is proposed in this paper to separate multiple sources from a mixture by designing a parallel dual generative adversarial Network (PDualGAN) that can build the relationship between a mixture and the corresponding multiple sources to achieve one-to-multiple cross-domain mapping. This algorithm can be applied to any mixing model such as the instantaneous mixing model and convolutive mixing model. In addition, datasets for one-to-multiple source separation are created which include the mixtures and corresponding sources for this study. Experiments were performed on four different datasets including both one-dimensional and two-dimensional signals. Experimental results show that the proposed algorithm can achieve high performance in peak signal-to-noise ratio (PSNR) and correlation, which outperforms state-of-the-art algorithms.
\end{abstract}


\IEEEpeerreviewmaketitle

\section{Introduction}
Single channel blind source separation (SCBSS) aims to separate multiple sources from a mixture collected by a sensor which has a variety of applications in speech \cite{1}, image \cite{2} and EEG signal processing \cite{3}, biomedical engineering\cite{4}\cite{5}\cite{6}, multi-source phase retrieval\cite{7}\cite{8}\cite{9}, and eddy current pulsed thermography \cite{10}, such as observe and analyze the anisotropies in the cosmic microwave background (CMB) radiation and recover the CMB component as accurately as possible from a noisy mixture. The challenge of SCBSS is that only a single-channel mixture is available, besides, there is no prior knowledge, and both the sources and mixing matrixes are unknown, so this problem is extremely underdetermined which leads to SCBSS being a very pathological and difficult problem.

Traditional methods like design the optimal filter (such as Wiener filter)\cite{12}\cite{13}, empirical mode decomposition and independent component analysis (EMD-ICA) \cite{14}\cite{15}, ensemble empirical mode decomposition and independent component analysis (EEMD-ICA) \cite{16}\cite{17}, non-negative matrix factorization (NMF)\cite{18}\cite{19} have been proposed to solve the problem of SCBSS. These methods are better for processing instantaneous mixing signals. However, filter based methods have high complexity and are not suitable for non-stationary random signals, the IMFs component obtained by the EMD and EEMD is prone to modal aliasing, ICA assumes that the signals are independent of each other, and NMF may ignore the phase. Recently, deep learning networks have seen increasing popularity as a solution to solving this problem. Auto-encoders (AEs)\cite{37} have been proposed as an approach to supervised source separation, but AEs rely on the assumption of an output probability density. Singing Voice Separation generative adversarial networks (SVSGAN) \cite{38} with a time-frequency masking function is proposed for separating voice sources. Wasserstein-GAN \cite{20} and two-stage approach \cite{39} are proposed for speech separation, which has a high computational cost. Evolving multi-resolution pooling convolutional neural
network (E-MRP-CNN) \cite{40} is proposed for monaural singing voice separation (MSVS). Synthesis-decomposition (S-D) \cite{21}, and single channel blind deconvolution algorithm based on deep convolution generating adversarial network (DCSS) \cite{22} are proposed for separating black and white images such as minist images from a mixture, but they are not suitable for processing color images. In addition, these methods have limited performance in separating multiple sources(more than three sources) and generalization, and they mainly focus on processing single structure signals.

In real life, the signal captured by sensor could be degraded by a number of conditions. For example, the electroencephalogram (EEG) signals are often mixed with interference signals of electrocardiogram (ECG), electromyography (EMG), and eye movement artifacts (EOG). The recorded remote sensing images are often obscured by clouds and fog. Due to the erosion of time, the loss of ancient Chinese characters often occurs in unearthed cultural relics and ancient books. Fingerprint images obtained from criminal investigation scenes often have multiple fingerprints overlapping each other. In a noisy acoustic environment, multiple speakers are talking simultaneously, which raises the challenge of separating multiple sources from a mixing signal. Although many methods have been proposed, separating multiple sources from a single channel mixture remains an open challenge.

The aim of this paper is to address the problem of single channel blind source separation of multiple signals that can be applied to both one-dimensional and two-dimensional signals based on instantaneous mixing model and convolutive mixing model, and improve its generalization performance.

Inspired by dual learning and GAN \cite{23} which can be used to build one-to-one mapping to achieve image-to-image translation, we design a parallel dual generative adversarial network (PDualGAN) to achieve one-to-multiple mapping, and formulate SCBSS as a data conversion problem in different domains where multiple sources are separated from a mixture in terms of the mapping between a mixture and the corresponding sources. Our novel contributions are as follows:

\begin{enumerate}[]
    \item {\textbf{Model.}} We have formulated a unified model for the instantaneous mixing model and convolutive mixing model.
	\item {\textbf{PDualGAN algorithm.}} A new algorithm is proposed by introducing parallel dual generative adversarial network (PDualGAN), where the model training is performed using the mixtures and corresponding multiple sources. The stability of the network and the quality of the generated data are improved by applying the Wasserstein GAN gradient penalty (WGAN-GP) loss function, thereby potentially enhancing source separation.

Our algorithm can be applied to both one-dimensional and two-dimensional signals. In addition, different mixtures are used as experimental data to test the effectiveness and generalization performance.
    \item {\textbf{Datasets.}} New datasets are created for this study which is composed of two parts: the mixtures and corresponding original sources. Each mixture is generated by using randomly generated mixing matrixes and multiple sources with different weights. The datasets created could be valuable for researchers working in the area of source separation and signal processing.

\end{enumerate}

This paper is organized as follows: Section II describes the background. Section III builds a one-to-multiple model for the SCBSS problem. Section IV presents our PDualGAN algorithm for the problem of SCBSS. Section V shows the experimental results. Section VI concludes the paper and draws future research directions.
\section{Background}
As we all know, training on datasets of smaller size while maintaining nearly the same performance would be very beneficial. `Less than one' shot learning \cite{24} can learn \emph{N} new classes given only \emph{M}  $\textless$ \emph{N} examples with the help of distance-weighted soft-label prototype k-Nearest Neighbors (SLaPkNN).

\textbf{Definition 1:} A hard label is a vector of length \emph{N}, indicating that the membership of a data point is exactly one of the \emph{N} classes.

\textbf{Definition 2:} A soft label is a vector representation of a data point's simultaneous membership to multiple classes.

\textbf{Definition 3:} Soft label prototype (SLaP) is a pair of vectors (\emph{X}, \emph{Y}), where \emph{X} is a feature (or position) vector, and \emph{Y} is an associated soft label.

\begin{figure}[!t]\centering
	\includegraphics[width=6cm]{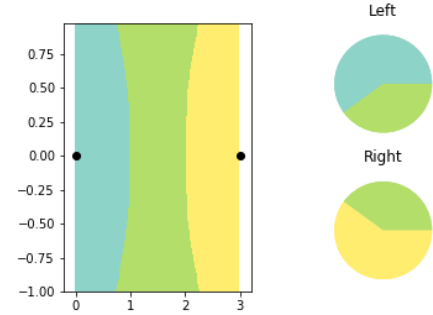}
	\caption{The soft label distribution of each prototype is illustrated by the pie charts.}\label{Fig_1}
\end{figure}

In classical \emph{k} nearest neighbor (kNN)\cite{25}, only hard labels are used. When a space is partitioned using the kNN classification rule, it is clear that at least \emph{n} points are required to separate \emph{n} classes. SLaPkNN is a combination of SLaP and kNN, and it generalizes the soft label prototype of kNN, which can be applied to separate \emph{N} classes effectively by using only \emph{M} soft label prototypes, where \emph{M}  $\textless$ \emph{N}. For example, a SLaPkNN classifier is fitted on two soft-label prototypes and partitions the space into three classes, as shown in Fig. \ref{Fig_1}.

\begin{figure*}[!t]\centering
	\includegraphics[width=10.5cm]{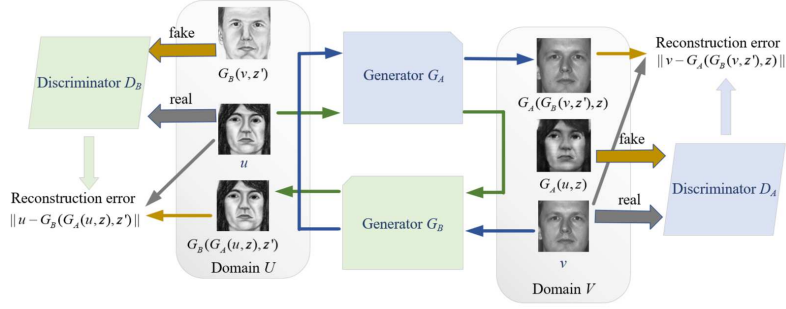}
	\caption{Working process of DualGAN for image-to-image translation. Image \emph{u} $\in$ \emph{U} is translated to domain \emph{V} using \emph{G}$_A$. Similarly, \emph{v} $\in$ \emph{V} is translated to \emph{U} using \emph{G}$_B$. }\label{Fig_2}
\end{figure*}

The `less than one' shot learning method requires labels in training, and therefore, it is not suitable for processing mixing signals. Inspired by this method, however, we develop an unsupervised mapping method to establish the relationship between the unlabeled and unpaired data.

A generative adversarial network (GAN) is a deep learning model and one of the most promising methods for unsupervised learning in recent years. The model produces good output through adversarial learning of the generator and the discriminator \cite{26}. However, the original GAN algorithm is not suitable for one-to-one conversion tasks between paired data.

Inspired by GAN and dual learning \cite{27} from natural language translation, a dual generative adversarial Network (DualGAN)\cite{23} is developed for one-to-one unlabeled data from two domains. The original GAN has the limitation that it can only learn to translate data from domain \emph{U} to those in domain \emph{V}, but the DualGAN can learn to invert the task.

Given two sets of unpaired and unlabeled data selected from domains \emph{U} and V, firstly, the task of the DualGAN \cite{23} is to learn a generator G$_A$: \emph{U} $\rightarrow$ \emph{V}, a mapping relation is built from \emph{u} $\in$ \emph{U} to \emph{v} $\in$ \emph{V}, secondly, the dual task is to train an inverse generator \emph{G}$_B$: \emph{V} $\rightarrow$ \emph{U}, and a mapping relation is built from \emph{v} $\in$ \emph{V} to \emph{u} $\in$ \emph{U}. This is realized with two GANs which have the same structure. The original GAN learns the generator \emph{G}$_A$ and discriminator \emph{D}$_A$ that discriminates between the fake and real data of domain V. Similarly, the dual GAN learns the generator \emph{G}$_B$ and a discriminator \emph{D}$_B$. The overall working process is shown in Fig. \ref{Fig_2}.
1
The \emph{u} $\in$ \emph{U} is translated to domain \emph{V} with \emph{G}$_A$. The fitting degree of \emph{G}$_A$(\emph{u}, \emph{z}) (\emph{z} is a random noise) is evaluated by \emph{D}$_A$,
Then, \emph{G}$_A$(\emph{u}, \emph{z}) is translated back to domain \emph{U} with \emph{G}$_B$, and output \emph{G}$_B$(\emph{G}$_A$(\emph{u}, \emph{z}), \emph{z}$'$) (\emph{z}$'$ is also random noise) as a reconstruction of \emph{u}. Similarly, \emph{v} $\in$ \emph{V} is translated to domain \emph{U} as \emph{G}$_B$(\emph{v}, \emph{z}$'$) with \emph{G}$_B$, and then reconstructed as G$_A$(G$_B$(\emph{v}, \emph{z}$'$), \emph{z}) with \emph{G}$_A$. The discriminator \emph{D}$_A$ is trained with \emph{v} as true data and \emph{G}$_A$(\emph{u}, \emph{z}) as fake data, however, \emph{D}$_B$ takes \emph{u} as true and \emph{G}$_B$(\emph{v}, \emph{z}$'$) as fake. The generators \emph{G}$_A$ and \emph{G}$_B$ are trained and optimized to output fake samples to cheat the corresponding discriminators \emph{D}$_A$ and \emph{D}$_B$, and to minimize the reconstruction error $\parallel\emph{v} - $\emph{G}$_A$(\emph{G}$_B$(\emph{v}, \emph{z}$'$),\emph{ z}))$\parallel$ and $\parallel$\emph{u} - \emph{G}$_B$(\emph{G}$_A$(\emph{u}, \emph{z}), \emph{z}$'$))$\parallel$.

The cross-entropy loss function of the original GAN \cite{26} is substituted by the loss function of Wasserstein GAN (WGAN)\cite{28}, which performs better in generator convergence and data quality, and in improving the stability of the network. The loss function applied in \emph{D}$_A$ and \emph{D}$_B$ can be written as
\begin{equation}
l_A^d\left( {u,v} \right) = {D_A}\left( {{G_A}\left( {u,z} \right)} \right) - {D_A}\left( v \right),\
\label{eqs:1}
\end{equation}
\begin{equation}
l_B^d\left( {u,v} \right) = {D_B}\left( {{G_B}\left( {v,z'} \right)} \right) - {D_B}\left( u \right),\
\label{eqs:2}
\end{equation}
where \emph{u} $\in$ \emph{U} and \emph{v} $\in$ \emph{V}.

The same loss function is applied in generators G$_A$ and G$_B$ as they have the same objective which adopts \emph{L}$_1$ distance to measure the reconstruction losses, it can be written as
\begin{equation}
\begin{aligned}
{l^g}\left( {u,v} \right) &= {\lambda _U}||u - {G_B}\left( {{G_A}\left( {u,z} \right),z'} \right)||\\
 &+ {\lambda _V}||v - {G_A}\left( {{G_B}\left( {v,z'} \right),z} \right)||\\
 &- D_B^{}\left( {{G_B}\left( {v,z'} \right)} \right) - {D_A}\left( {{G_A}\left( {u,z} \right)} \right)
\label{eqs:3}
\end{aligned}
\end{equation}
where $\lambda_U$ and  $\lambda_V$ are both constant parameters, which are typically set to a value within [100, 1000].

Clearly, by training DualGAN, unlabeled and unpaired data can be converted from \emph{U} to the corresponding data in \emph{V} because the data of the two domains have some similar characteristics. Inspired by DualGAN, single channel blind source separation can be realized by converting the mixtures to the corresponding multiple sources which have similar characteristics. In this paper, we develop a PDualGAN to train multiple DualGAN simultaneously, and convert the mixtures to corresponding multiple sources using the PDualGAN algorithm.

\section{Mathematical Model}
The instantaneous mixing model and the convolutive mixing model used in single channel blind source separation are given below.

\begin{enumerate}[]
    \item {\textbf{Instantaneous mixing model.}}

For instantaneous mixing model, the observed mixture \emph{x}(\emph{t}) can be defined as
\begin{equation}
x(t) = \sum\limits_{i = 1}^N {{a_i}\left( t \right)} {s_i}\left( t \right),\
\label{eqs:4}
\end{equation}
where \emph{x}(\emph{t}) is a observed mixture, \emph{s}$_i$(\emph{t}) is the \emph{i}th source, \emph{a}$_i$(\emph{t}) is the \emph{i}th mixing matrix. \emph{i} =1, 2, ..., \emph{N}.

	\item {\textbf{Convolutive mixing model.}}

For convolutive mixing model, the observed mixture \emph{x}(\emph{t}) can be defined as
\begin{equation}
x(t) = \sum\limits_{i = 1}^N {{a_i}\left( t \right)}*{s_i}\left( t \right),\
\label{eqs:5}
\end{equation}
where \emph{x}(\emph{t}) is a observed mixture, \emph{s}$_i$(\emph{t}) is the \emph{i}th source, \emph{a}$_i$(\emph{t}) is the \emph{i}th mixing matrix. \emph{i} =1, 2, ..., \emph{N}, * denotes a convolution operation which can be defined as
\begin{equation}
{a_i}\left( t \right) * {s_i}\left( t \right) = \sum\limits_{i = 1}^N {{a_i}\left( {t - v} \right)} {s_i}\left( v \right),\
\label{eqs:6}
\end{equation}
where \emph{t} is the amount of displacement of $\emph{a}$$(-\emph{v})$. So \eqref{eqs:5} can be described as
\begin{equation}
x\left( t \right) = \sum\limits_{i = 1}^N {{a_i}(t - v)} {s_i}(v).\
\label{eqs:7}
\end{equation}
\end{enumerate}

We can see from \eqref{eqs:4} and \eqref{eqs:7} that both the instantaneous mixing model and the convolutive mixing model can be regarded as a form of matrix multiplication, and a mixing signal corresponds to multiple sources. The aim of this paper is to find the one-to-multiple mapping relationship between the mixtures and the corresponding multiple sources.
\section{PDualGAN Algorithm}
\begin{figure*}[!t]\centering
	\includegraphics[width=14cm]{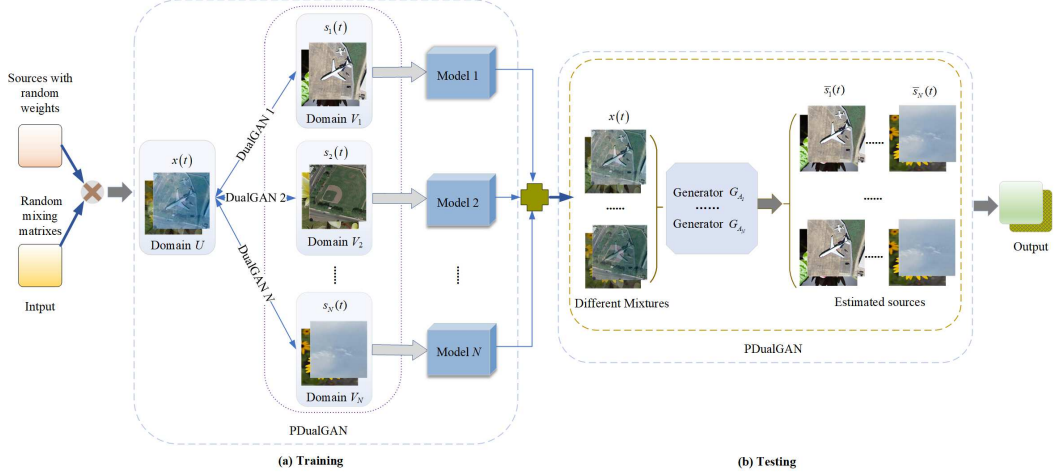}
	\caption{The architecture of PDualGAN algorithm based on the instantaneous mixing model and convolutive mixing model, including the training of PDualGAN (a), where \emph{N} DualGAN are trained simultaneously, which learn the mapping between the mixture \emph{x}(\emph{t}) $\in$ \emph{U} and the corresponding sources \emph{s}$_1$(\emph{t}) $\in$ \emph{V$_1$}, \emph{s}$_2$(\emph{t}) $\in$ \emph{V$_2$}, ..., \emph{s}$_N$(\emph{t}) $\in$ \emph{V$_N$} sequentially, each DualGAN learns one of these mappings. The testing of PDualGAN (b), where mixtures are generated by random mixing matrixes and same sources of different weights (different from the mixing weights of training). \emph{x}(\emph{t}) are used as the input for estimating the \emph{N} sources ${\bar s}$$_1$(\emph{t}), ${\bar s}$$_2$(\emph{t}), ..., ${\bar s}_N$(\emph{t}) as the output.}\label{Fig_3}
\end{figure*}
\begin{figure*}[!t]\centering
	\includegraphics[width=14cm]{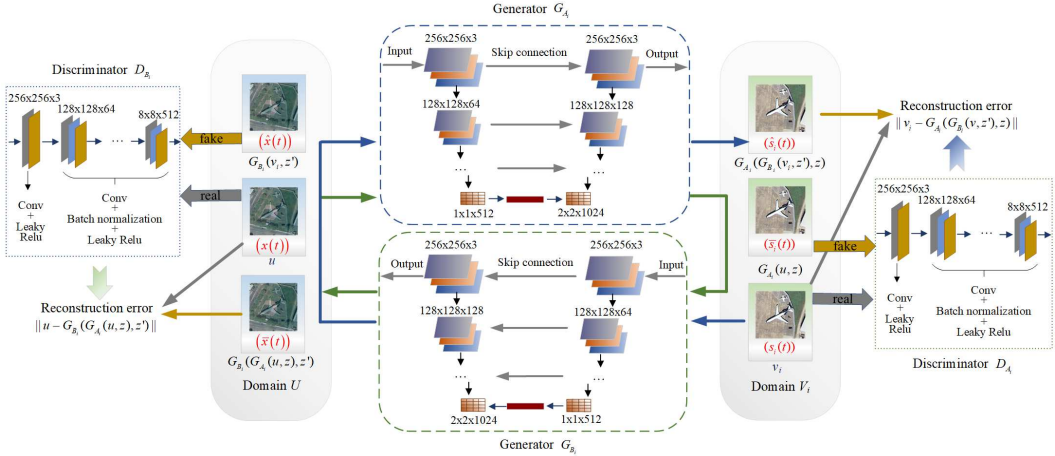}
	\caption{Working process of each DualGAN. ${\bar s}$$_i$(\emph{t})$\in$ \emph{V$_i$} is the fake data generated by generator \emph{G}$_{A_i}$ from the mixture \emph{x}(\emph{t})$\in$ \emph{U}, ${\bar x}$(\emph{t})$\in$ \emph{U} is reverse generated data by generator \emph{G}$_{B_i}$. Similarly, ${\hat x}$(\emph{t})$\in$ \emph{U} is the fake data generated by generator \emph{G}$_{B_i}$ from the sources \emph{s$_i$}(\emph{t})$\in$ \emph{V$_i$}, ${\hat s}$$_i$(\emph{t})$\in$ \emph{V$_i$} is reverse generated data by generator \emph{G}$_{A_i}$. \emph{x}(\emph{t}) (i.e. \emph{u}), \emph{s}$_i$(\emph{t}) (i.e. \emph{v}$_i$), ${\bar x}$(\emph{t}) (i.e. \emph{G}$_{B_i}$(\emph{G}$_{A_i}$(\emph{u}, \emph{z}), \emph{z}$'$)), ${\bar s}$$_i$(\emph{t}) (i.e. \emph{G}$_{A_i}$(\emph{u}, \emph{z})), ${\hat x}$(\emph{t}) (i.e. \emph{G}$_{B_i}$(\emph{v}$_i$, \emph{z}$'$)), ${\hat s}$$_i$(\emph{t}) (i.e. \emph{G}$_{A_i}$(\emph{G}$_{B_i}$(\emph{v}$'$, \emph{z}$'$), \emph{z})) }\label{Fig_4}
\end{figure*}

In this section, we present a PDualGAN algorithm to address the problem of separating multiple sources from a mixture based on \eqref{eqs:4} and \eqref{eqs:7} which is applied to one-dimensional and two-dimensional signals. As shown in Fig. \ref{Fig_3}, we transform the problem of SCBSS into a data conversion problem in different domains by using PDualGAN to train \emph{N} DualGAN simultaneously. The first step is to train the PDualGAN with the mixtures (instantaneous mixtures or convolutive mixtures) and corresponding sources. Mixtures are generated by using randomly generated mixing matrixes and multiple sources with different weights (i.e. different sources have different proportions), and the mixing matrixes satisfy the Gaussian or normal distribution. The second step is to test the effect of the trained model with different mixtures of different weights of the same sources.

In our algorithm, no specific domain knowledge or pre-trained domain representation is needed, but the mapping relationship between unpaired signals is searched. The reconstruction error measures the disparity between the original signals and the reconstructed signals. PDualGAN contains \emph{N} DualGANs, which are of the same structure as shown in Fig. \ref{Fig_4}. The mixing signals are sampled from \emph{U} and the corresponding sources are sampled from \emph{V}$_1$, \emph{V}$_2$, ..., \emph{V}$_N$ respectively, The primary task of our PDualGAN is to build the one-to-multiple mapping from \emph{x}(\emph{t})$\in$ \emph{U} to \emph{s}$_1$(\emph{t}), \emph{s}$_2$(\emph{t}), ..., \emph{s}$_N$(\emph{t}) $\in$ \emph{V$_1$}, \emph{V$_2$},..., \emph{V$_N$}.

\textbf{Network configuration.}
Our PDualGAN including \emph{N} DualGANs of the same structure, each DualGAN has identical network architecture for \emph{G}$_{A_i}$ and \emph{G}$_{B_i}$(\emph{G}$_{A_i}$ and \emph{G}$_{B_i}$ are the generators of the \emph{i}th DualGAN, \emph{i}=1, 2, ..., \emph{N}). The generator has the equal number of upsampling and downsampling layers, with skip connections between them, forming a U-shaped net \cite{29}\cite{30}, and such a structure enables low-level information to be shared between the input and output. In addition, \emph{z} and \emph{z}$'$ are provided only in the form of dropout and applied to multiple layers of generators at both training and testing phases, but they are not explicitly provided. For discriminators, the Markovian PatchGAN architecture \cite{31} is applied, which is effective in capturing local high-frequency features (such as texture in an image), and its effectiveness has been verified on various conversion tasks \cite{27}. Furthermore, it requires fewer parameters, and therefore runs faster than conventional GAN. This scheme will be used for our signal separation tasks.

We train the discriminators n$_{critic}$ steps, then one step on generators. The number of critic iterations per generator iteration n$_{critic}$ can be set to 2-4, $\lambda$$_U$ and $\lambda$$_{V_i}$ are all set to 1000, an initial learning rate is set to 0.00005, and the batch size is assigned as 1. The clipping parameter is set in [0.01, 0.1], and the epoch is set to 20000.

\textbf{Training.}
As the momentum-based methods (such as Adam) would occasionally lead to instability, we use mini-batch stochastic gradient descent (SGD) and apply the RMSProp solver which is known to perform well on highly nonstationary problems \cite{28}. The sigmoid cross-entropy loss of the traditional GAN is locally saturated and may cause the gradient to disappear. However,  Wasserstein GAN gradient penalty (WGAN-GP)\cite{28} loss is differentiable almost everywhere, resulting in a better discriminator which can provide more reliable gradient information.

As shown in Fig. \ref{Fig_3} (a) and Fig. \ref{Fig_4}, for a mixture, a generator \emph{G}$_{A_i}$: \emph{U} $\rightarrow$ \emph{V}$_i$ in the \emph{i}th DualGAN is learned by mapping the mixture \emph{u} to a generated corresponding source \emph{G}$_{A_i}$(\emph{u}, \emph{z}), while the dual task is to train an inverse generator \emph{G}$_{B_i}$: \emph{V}$_i$ $\rightarrow$ \emph{U} that maps a generated source \emph{G}$_{A_i}$(\emph{u}, \emph{z}) to a generated mixture \emph{G}$_{B_i}$(\emph{G}$_{A_i}$(\emph{u}, \emph{z}), \emph{z}$'$), where \emph{z} and \emph{z}$'$ are random noise signals. \emph{N} corresponding sources are simultaneously generated by \emph{N} DualGAN (i.e. $i=1, 2, ...,  \emph{N}$) from the same mixture with the identical structure.

For an original source, a generator \emph{G}$_{B_i}$: \emph{V}$_i$ $\rightarrow$ \emph{U} in the \emph{i}th DualGAN is learned by mapping the source \emph{v}$_i$ to a generated mixture \emph{G}$_{B_i}$(\emph{v}$_i$, \emph{z}$'$), while the dual task is to train an inverse generator \emph{G}$_{A_i}$: \emph{U} $\rightarrow$ \emph{V}$_i$ that maps a generated mixture \emph{G}$_{B_i}$(\emph{v}$_i$, \emph{z}$'$) to a generated source \emph{G}$_{A_i}$(\emph{G}$_{B_i}$(\emph{v}$_i$, \emph{z}$'$), \emph{z}), where \emph{z} and \emph{z}$'$ are random noises. \emph{N} different sources generate the same mixture by \emph{N} DualGAN (i.e. $i=1, 2, ..., \emph{N}$) with the identical structure.

The discriminator \emph{D}$_{A_i}$ discriminates the real source \emph{v}$_i$ of domain \emph{V}$_i$ and the fake outputs \emph{G}$_{A_i}$(\emph{u}, \emph{z}). The discriminator \emph{D}$_{B_i}$ discriminates the real mixture \emph{u} of domain \emph{U} and the fake outputs \emph{G}$_{B_i}$(\emph{v}$_i$, \emph{z}$'$)(\emph{D}$_{A_i}$ and \emph{D}$_{B_i}$ are the discriminators of the \emph{i}th DualGAN, \emph{i}=1, 2, ..., \emph{N}).

The same loss function is applied in each DualGAN for generators \emph{G}$_{A_i}$ and \emph{G}$_{B_i}$ which is defined as
\begin{equation}
\begin{aligned}
{l^{{G_i}}}\left( {u,{v_i}} \right) &= {\lambda _U}||u - {G_{{B_i}}}\left( {{G_{{A_i}}}\left( {u,z} \right),z'} \right)||\\
 &+ {\lambda _{V_i}}||{v_i} - {G_{{A_i}}}\left( {{G_{{B_i}}}\left( {{v_i},z'} \right),z} \right)||\\
 &- D_{{B_i}}^{}\left( {{G_{{B_i}}}\left( {{v_i},z'} \right)} \right) - {D_{{A_i}}}\left( {{G_{{A_i}}}\left( {u,z} \right)} \right),
\label{eqs:8}
\end{aligned}
\end{equation}
where \emph{u} $\in$ \emph{U}, \emph{v}$_i$ $\in$ \emph{V}$_i$, and $\lambda$$_U$,
$\lambda$$_{V_i}$ are two constant parameters which are set to a value within [100.0, 1000.0].

The corresponding loss functions applied in \emph{D}$_{A_i}$ and \emph{D}$_{B_i}$ are defined as:
\begin{equation}
l_{{A_i}}^D\left( {u,{v_i}} \right) = {D_{{A_i}}}\left( {{G_{{A_i}}}\left( {u,z} \right)} \right) - {D_{{A_i}}}\left( {{v_i}} \right),\
\label{eqs:9}
\end{equation}
\begin{equation}
l_{{B_i}}^D\left( {u,{v_i}} \right) = {D_{{B_i}}}\left( {{G_{{B_i}}}\left( {{v_i},z'} \right)} \right) - {D_{{B_i}}}\left( u \right),\
\label{eqs:10}
\end{equation}
where \emph{u} $\in$ \emph{U}, and \emph{v}$_i$ $\in$ \emph{V}$_i$.

The one-to-multiple mapping between a mixture \emph{u} $\in$ \emph{U} and \emph{N} sources \emph{v}$_1$, \emph{v}$_2$, ..., \emph{v}$_N$ $\in$ \emph{V}$_1$, \emph{V}$_2$, ..., \emph{V}$_N$, respectively, is built by training the proposed PDualGAN.

\textbf{Testing.}
As shown in Fig. \ref{Fig_3} (b), after training the proposed PDualGAN, we save the trained model and parameters which are then used in the test stage. The mixtures are passed through the trained model and converted to corresponding multiple sources to achieve separation of multiple sources from the mixture.

\begin{algorithm}
\renewcommand{\algorithmicrequire}{\textbf{Input:}}
\renewcommand{\algorithmicensure}{\textbf{Iteration}}
\newcommand{\LASTCON}{\item[\algorithmiclastcon]}
\newcommand{\algorithmiclastcon}{\textbf{Output:}}
\footnotesize
\caption{PDualGAN Algorithm}
\label{alg1}
\begin{algorithmic}[1]
    \REQUIRE The mixture \emph{u} $\in$ \emph{U}, original sources \emph{v}$_i$ $\in$ \emph{V}$_i$, \emph{i} = 1, 2, ..., \emph{N}, $\lambda$$_U$, $\lambda$$_{V_i}$,  initial learning
rate,  batch size, clipping parameter.
    \LASTCON Estimated sources \emph{v}$_i$ $\in$ \emph{V}$_i$, \emph{i} = 1, 2, ..., \emph{N}.
    \STATE Each mixture \emph{X}(\emph{t}) is mapped to \emph{N} corresponding sources \emph{s}$_1$(\emph{t}), \emph{s}$_2$(\emph{t}), ..., \emph{s}$_N$(\emph{t}) by \emph{N} generators \emph{G}$_{A_1}$, \emph{G}$_{A_2}$, ...,  \emph{G}$_{A_N}$, and \emph{N} original sources are mapped to the same mixture by \emph{N} generators \emph{G}$_{B_1}$, \emph{G}$_{B_2}$, ..., \emph{G}$_{B_N}$ simultaneously. The parallel results are saved.

    \STATE Optimize the loss function of generators \emph{G}$_{A_i}$ and \emph{G}$_{B_i}$ in each DualGAN.
\[\begin{array}{l}
{l^{{G_i}}}\left( {u,{v_i}} \right) = {\lambda _U}||u - {G_{{B_i}}}\left( {{G_{{A_i}}}\left( {u,z} \right),z'} \right)||\\
 + {\lambda _{{V_i}}}||{v_i} - {G_{{A_i}}}\left( {{G_{{B_i}}}\left( {{v_i},z'} \right),z} \right)||\\
 - D_{{B_i}}^{}\left( {{G_{{B_i}}}\left( {{v_i},z'} \right)} \right) - {D_{{A_i}}}\left( {{G_{{A_i}}}\left( {u,z} \right)} \right)
\end{array}\]
    \STATE Optimize the loss function of discriminators \emph{D}$_{A_i}$ and \emph{D}$_{B_i}$ in each DualGAN.
\[l_{{A_i}}^D\left( {u,{v_i}} \right) = {D_{{A_i}}}\left( {{G_{{A_i}}}\left( {u,z} \right)} \right) - {D_{{A_i}}}\left( {{v_i}} \right)\]
\[l_{{B_i}}^D\left( {u,{v_i}} \right) = {D_{{B_i}}}\left( {{G_{{B_i}}}\left( {{v_i},z'} \right)} \right) - {D_{{B_i}}}\left( u \right)\]
    \STATE end.
\end{algorithmic}
\end{algorithm}

\section{Numerical Experiments}
In this section, the experiment is conducted to demonstrate the performance of the proposed PDualGAN algorithm, both one-dimensional and two-dimensional signals are used in the experiment. Besides, take the separation of four sources from a mixture as an example, we compare the PDualGAN algorithm with state-of-the-art baseline algorithms.

\textbf{Experimental data.}
We use four kinds of datasets for the experiment: the NWPU-occlusion image datasets, the ancient Chinese character occlusion (ACC-occlusion) datasets, the speech datasets, and the EEG datasets. We select a total of 2000 original sources, each dataset includes 500 sources, mixtures are generated by using randomly generated mixing matrixes and four sources with different weights. Each dataset includes two types: (1) instantaneous mixtures and corresponding sources; (2) convolutive mixtures and corresponding sources.

Instantaneous mixtures are generated by using randomly generated mixing matrixes and four sources with different weights. The convolutional mixed signal is the product of the four sources with different weights and the randomly generated mixing matrixes after a reversal and shift. The mixing matrixes satisfy the Gaussian or normal distribution.

[NWPU-occlusion image datasets] A NWPU-occlusion image dataset including the original images selected from the NWPU-RESISC dataset\footnote{https://hyper.ai/datasets/5449} containing 45 categories and cloud and fog occlusion images captured by us from Google and Baidu search engines. Each mixing image is generated by mixing two NWPU-RESISC images and two cloud and fog occlusion images in different random weights.

[ACC-occlusion datasets] An ACC-occlusion dataset including the ancient Chinese character images and occlusion images was built. Each mixing image is obtained by mixing one ancient Chinese character image and three occlusion images in different random weights.

[Speech datasets] The speech signals are selected from the THCHS-30 dataset\footnote{http://www.openslr.org/resources/18/data} with each sentence containing 16384 points randomly clipped and then normalized. Each mixing signal is obtained by mixing four randomly selected speech signals in different random weights. The THCHS-30 dataset was recorded by a single carbon microphone in a quiet office environment, where the signals were sampled at 16 kHz and quantized in 16 bits. Most of the speakers are college students who can speak fluent Mandarin. In total, 1000 recordings are taken, which have a total duration of more than 30 hours.

[EEG datasets] The EEG acquisition equipment used in the experiment is the EEG and evoked potential meter (model NCERP-T-240) of Shanghai Nuocheng Electric Co., Ltd. We use 24-channel silver-plated electrodes, and the electrode placement position adopts the international standard 10/20. An attention device using the TGAM chip was developed to test the attention of the participants, where a pre-defined threshold \emph{p} can be used to detect the attention of the participant. If it is higher than the pre-defined threshold \emph{p} (e.g. \emph{p} $\textgreater$ 60), it indicates that the participant has a good focus on the auditory stimuli, and at this time, EEG signals can be collected by the device.

\begin{table*}[htbp]
    \renewcommand{\arraystretch}{1.5}
    \renewcommand\tabcolsep{7pt}
 \caption{The performence of our algorithm as compared with state-of-the-art algorithms on four different datasets.}
 \centering
 \label{tab1}
  \begin{tabular}{cccccccccccccccccc}
   \hline\hline \\[-3mm]
             \multirow{3}{*}{Algorithm} & \multicolumn{6}{c}{Datasets} \\
                                        & \multicolumn{2}{c}{NWPU-occlusion} & \multicolumn{2}{c}{ACC-occlusion} & \multicolumn{1}{c}{Speech} & \multicolumn{1}{c}{EEG}\\
                                        & PSNR & Correlation & PSNR  & Correlation & Correlation & Correlation\\ \hline
   \multicolumn{1}{l}{Instantaneous model based algorithms}      \\ \hline
   \multicolumn{1}{l}{EMD-ICA}  & 7.91    & 0.2198               & 7.73  & 0.1583          & 0.1779    & 0.1113    \\ \hline
   \multicolumn{1}{l}{EEMD-PCA-ICA}  & 8.36    & 0.2730        & 8.84  & 0.2947          & 0.1188    & 0.1332    \\ \hline
   \multicolumn{1}{l}{CEEDMAN-ICA}  & 7.50    & 0.1666               & 7.61  & 0.2015          & 0.1349    & 0.1564    \\ \hline
   \multicolumn{1}{l}{SSA-ICA}  & 8.92    & 0.2568               & 8.85  & 0.2768          & 0.1997    & 0.1173    \\ \hline
   \multicolumn{1}{l}{Gaussian WGAN}  & 7.75 & 0.2961            & 7.92  &  0.2876         &  0.3547  &  0.1451     \\ \hline
   \multicolumn{1}{l}{PDualGAN}  & 20.15    & 0.7212               & 24.98  & 0.7819          & 0.7413    & 0.6862    \\ \hline \hline
   \multicolumn{1}{l}{Convolutive model based algorithms}    \\ \hline
   \multicolumn{1}{l}{Conjugate Gradieng}  & 4.35    & 0.1095               & 5.32  & 0.1152          & 0.1339    & 0.1812    \\ \hline
   \multicolumn{1}{l}{IRLS}  & 4.04    & 0.1113               & 4.85  & 0.1372          & 0.1538    & 0.1430    \\ \hline
   \multicolumn{1}{l}{S-D}  & 11.81    & 0.5259               & 13.59  & 0.7049          & 0.4520    & 0.3063    \\ \hline
   \multicolumn{1}{l}{E-MRP-CNN}  &  9.15   &  0.4313              & 9.52  &  0.4493         &  0.6285   &  0.6031   \\ \hline
   \multicolumn{1}{l}{PDualGAN}  & 19.38    & 0.7026               & 22.76  & 0.7792          & 0.7422    & 0.6545    \\[0.5ex]
   \hline\hline
  \end{tabular}
\end{table*}

A total of 50 participants are included in the EEG signal collection, of which 25 are males and 25 are females, aged between 20-40. All participants are either teachers or students from Taiyuan University of Science and Technology who are in good health and meet the requirements for EEG collection experiments.

We select the EEG signals according to the positions of the EEG electrodes of the temporal and frontal lobes and pre-processing them to remove artifacts such as ECG and oculus, then we randomly clipped 16384 points in each channel and normalize and convert the selected EEG signals into a multi-dimensional matrix to construct an EEG dataset. Each mixture is obtained by mixing four original EEG signals from the collected dataset in different random weights.

We conducted experiments on these four datasets according to the two mixing modes. For one-dimensional and two-dimensional signals, we show the test results of one of instantaneous mixtures and convolutive mixtures respectively (i.e. for two-dimensional signals, We show the separation result of the instantaneous mixtures on the NWPU-occlusion image datasets, and the separation result of the convolutive mixtures on the ACC-occlusion datasets; for one-dimensional signals, We show the separation result of the instantaneous mixtures on the EEG datasets, and the separation result of the convolutive mixtures on the Speech datasets).

Our PDualGAN algorithm is compared with the state-of-the-art algorithms on these four kinds of signals. The four kinds of datasets are used to train our network, all the datasets are pre-processed, each mixture corresponds to four sources, which as the input data of the PDualGAN network. Finally, different mixtures of four sources of different weights are selected as test signals to demonstrate the effectiveness of the proposed algorithm.

\textbf{Baseline method} We compare our PDualGAN algorithm with baseline algorithms for both instantaneous and convolutive mixing model based algorithms in separating four sources from a mixture. Instantaneous mixing model based algorithms including EMD-ICA (empirical mode decomposition and independent component analysis)\cite{15}, EEMD-PCA-ICA (ensemble empirical mode decomposition and principal component analysis and independent component analysis) \cite{16}, CEEDMAN-ICA (complete ensemble empirical mode decomposition with adaptive noise and independent component analysis) \cite{32}, SSA-ICA\footnote{https://download.csdn.net/download/qq\_39065549/12318104}(singular spectrum analysis and independent component analysis) \cite{33} and Gaussian WGAN\footnote{https://github.com/ycemsubakan/sourceseparation\_misc} \cite{20} algorithms. Convolutive mixing model based algorithms including Conjugate Gradient\footnote{http://groups.csail.mit.edu/graphics/CodedAperture\label{web}} \cite{36}, IRLS\footnote{http://groups.csail.mit.edu/graphics/CodedAperture\label{web}} \cite{36}, S-D (synthesis-decomposition)\footnote{https://github.com/qiuqiangkong/gan separation deconvolution} \cite{21}, and E-MRP-CNN \cite{40}\footnote{https://github.com/tuxzz/emrpcnn\_pub} algorithms.

The implement of EMD-ICA, EEMD-PCA-ICA, and CEEDMAN-ICA are downloaded from https://so.csdn.net/so/search?q=CEEMDAN\_V00\&t=doc\&u=\&urw and invoke emd.m, eemd.m and ceemdan.m respectively, then processed by FastICA to find the independent components by maximizing the statistical independence of the dimensionality reduction IMFs, PCA is used to extract the main energy in EEMD-PCA-ICA algorithm. For CEEDMAN-ICA algorithm, the data needs to be preprocessed and converted to mat format. The implement of Gaussian WGAN is achieved by running the main$\_$timit.py file for single source separation, and the data needs to be preprocessed and converted to wav format. For SSA-ICA and S-D algorithms, we adjusted part of the program code so that it can input and output four sources. The codes of EMD-ICA, EEMD-PCA-ICA, CEEDMAN-ICA, SSA-ICA, S-D, Gaussian WGAN and E-MRP-CNN algorithms cannot be used directly to process color images, so we process the R, G, and B channels separately. For E-MRP-CNN algorithm, we use the mixtures and the corresponding four sources selected from our datasets as the input data of this algorithm for training and testing, which are preprocessed and converted to wav format. The E-MRP-CNN algorithm only estimates two channels, one as the vocal source, and the other as the interference source composed of the remaining sources, so the average value is calculated according to the results of these two channels.

For the experiments on the NWPU-occlusion image datasets and ACC-occlusion datasets, the performance of the proposed PDualGAN algorithm can be evaluated by the peak signal-to-noise ratio (PSNR) \cite{34} and signal correlation \cite{35}. For the experiments on the speech datasets and EEG datasets, the performance of the proposed PDualGAN algorithm can be evaluated by signal correlation \cite{35}.

The PSNR can be defined as
\begin{equation}
{\mathop{\rm PSNR}\nolimits}  = 10{\log _{10}}\left( {\frac{{{{{\mathop{\rm MAX}\nolimits} }_I}^2}}{{{\mathop{\rm MSE}\nolimits} }}} \right),
\label{eqs:9}
\end{equation}
where MSE is the mean square error between two images with a size of \emph{w} $\times$ \emph{v}. MAX represents the maximum value of an image without noise. The MSE of images \emph{M} and \emph{H} can be written as
\begin{equation}
{\mathop{\rm MSE}\nolimits} (M,H) = \frac{1}{{wv}}{\sum\limits_{i = 0}^{w - 1} {\sum\limits_{j = 0}^{v - 1} {(M(i,j) - H(i,j)} } )^2}.
\label{eqs:10}
\end{equation}

The correlation coefficient of signals \emph{X} and \emph{Y} can be expressed as
\begin{equation}
r = \frac{{\sum\limits_{i = 1}^N {\left( {{X_i} - \bar X} \right)\left( {{Y_i} - \bar Y} \right)} }}{{\sqrt {\sum\limits_{i = 1}^N {{{\left( {{X_i} - \bar X} \right)}^2}} } \sqrt {\sum\limits_{i = 1}^N {{{\left( {{Y_i} - \bar Y} \right)}^2}} } }}.
\label{eqs:11}
\end{equation}

A higher PSNR value and signal correlation means a better effect of separating multiple sources, and a more closer the estimated source to the real source.

The facilities applied to perform the experiments include Intel I9-10900X 13.7 GHz CPU, 2*NVIDIA RTX 8000 Graphics Card and 6*32 GB memory. Gaussian WGAN, S-D, E-MRP-CNN and the proposed PDualGAN algorithms are implemented on this facilities. EMD-ICA\cite{15}, EEMD-PCA-ICA \cite{16}, CEEDMAN-ICA \cite{32}, SSA-ICA \cite{33} and Conjugate Gradient \cite{36}, IRLS \cite{36} algorithms are implemented in MATLAB. TABLE \ref{tab1} shows the results of separating four sources of our algorithm as compared with the state-of-the-art algorithms in mean values of PSNR and signal correlation on four different datasets. Clearly, the proposed PDualGAN algorithm has better performance than the EMD-ICA, EEMD-PCA-ICA, CEEDMAN-ICA, SSA-ICA, Gaussian WGAN, Conjugate Gradient \cite{36}, IRLS \cite{36}, S-D \cite{21}, and E-MRP-CNN \cite{40} algorithms.
\begin{figure}[htbp]\centering
	\includegraphics[width=7.8cm]{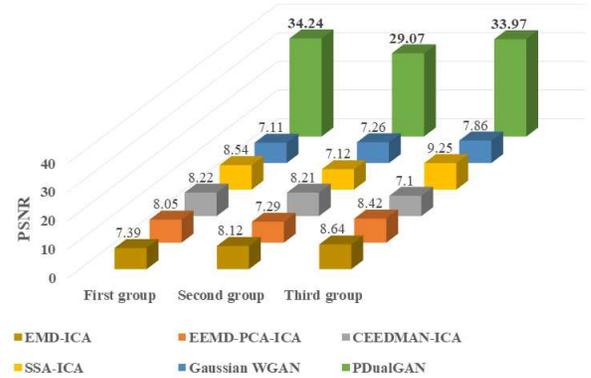}
	\caption{Average PSNR of the proposed PDualGAN as compared with five state-of-the-art instantaneous mixing model based algorithms.}\label{Fig_6}
\end{figure}

\begin{figure*}[!t]\centering
	\includegraphics[width=12cm]{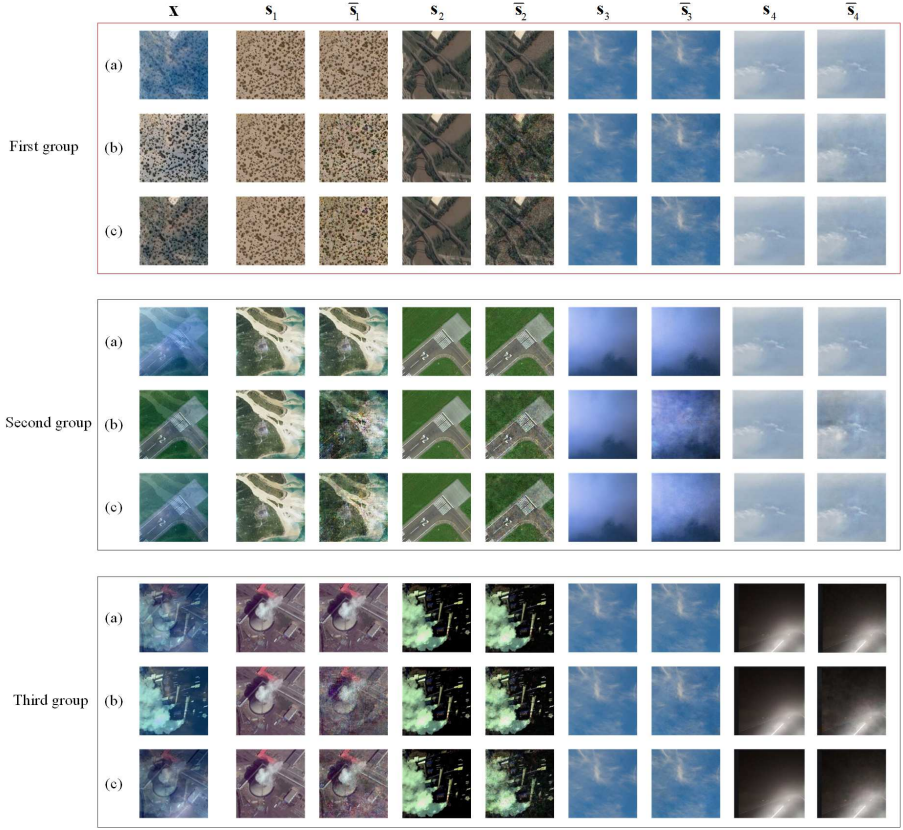}
	\caption{Testing results of PDualGAN approach for instantaneous mixing NWPU-occlusion images. Each group includes three different mixtures ((a), (b) and (c)) of four same images of different random weights (different from the mixtures of training). The $\mathbf{X}$ represents the mixture, $\mathbf{s}$$_1$, $\mathbf{s}$$_2$, $\mathbf{s}$$_3$, and $\mathbf{s}$$_4$ represent the four original ground-truth sources, $\mathbf{\bar s}$${_1}$, $\mathbf{\bar s}$${_2}$, $\mathbf{\bar s}$${_3}$, and $\mathbf{\bar s}$${_4}$ represent the corresponding estimated sources. }\label{Fig_5}
\end{figure*}

\begin{table*}[htbp]
    \renewcommand{\arraystretch}{1.4}
    \renewcommand\tabcolsep{7pt}
 \caption{Average correlation of the proposed PDualGAN as compared with five state-of-the-art convolutive mixing algorithms.}
 \centering
 \label{tab2}
  \begin{tabular}{ccccccccccccccccccc}
   \hline\hline \\[-1mm]
             \multirow{2}{*}{Algorithm}
                                        & \multicolumn{3}{c}{First group} & \multicolumn{3}{c}{Second group} & \multicolumn{3}{c}{Third group}\\
                                        & (a) & (b) & (c) & (a) & (b) & (c) & (a) & (b) & (c)  \\ \hline
   \multicolumn{1}{l}{EMD-ICA} & 0.2077 & 0.2098  &0.2055  &  0.2132  & 0.2188  & 0.2153 & 0.2754 & 0.2539 & 0.2675        \\ \hline
   \multicolumn{1}{l}{EEMD-PCA-ICA}& 0.2769 & 0.2688  &0.2805  &  0.1432  & 0.1179  & 0.1543 & 0.3134 & 0.3033 & 0.2955 \\ \hline
   \multicolumn{1}{l}{CEEDMAN-ICA}& 0.1479 & 0.1598  &0.1455  &  0.2132  & 0.2136  & 0.2203 & 0.3754 & 0.1839 & 0.2075 \\ \hline
   \multicolumn{1}{l}{SSA-ICA}& 0.3112 & 0.2526  &0.3670  &  0.2122  & 0.3433  & 0.1210 & 0.2714 & 0.2134 & 0.2865 \\ \hline
   \multicolumn{1}{l}{Gaussian WGAN}& 0.1689 & 0.3754 & 0.1813 & 0.2019 & 0.1983 & 0.1927 & 0.2154 & 0.2282 & 0.2396 \\ \hline
   \multicolumn{1}{l}{PDualGAN}& 0.9907 & 0.9201  &0.9484  &  0.9948  & 0.8443  & 0.9587 & 0.9784 & 0.9471 & 0.9491
   \\[0.5ex]
   \hline\hline
  \end{tabular}
\end{table*}

\subsection{PDualGAN for NWPU-occlusion images}

In the first set of simulations, we evaluate the separating performance of the proposed PDualGAN algorithm described in Algorithm 1 for instantaneous mixtures on the NWPU-occlusion image datasets.

In the testing, we randomly select three groups of the instantaneous mixtures and corresponding sources from NWPU-occlusion datasets, each group shows the separating results from three different mixtures ((a), (b), (c) in Fig. \ref{Fig_5}), and each mixture is instantaneously mixing generated by randomly selected mixing matrixes and four same images with different random weights. Fig. \ref{Fig_5} shows the test results of separating four images from instantaneous mixtures. Fig. \ref{Fig_6} shows the comparison results of average PSNR values obtained by the proposed PDualGAN algorithm and EMD-ICA, EEMD-PCA-ICA, CEEDMAN-ICA, SSA-ICA, Gaussian WGAN instantaneous model based algorithms. TABLE \ref{tab2} shows the comparison results of average correlation obtained by the proposed PDualGAN algorithm and these state-of-the-art algorithms.
\begin{figure*}[htbp]\centering
	\includegraphics[width=12cm]{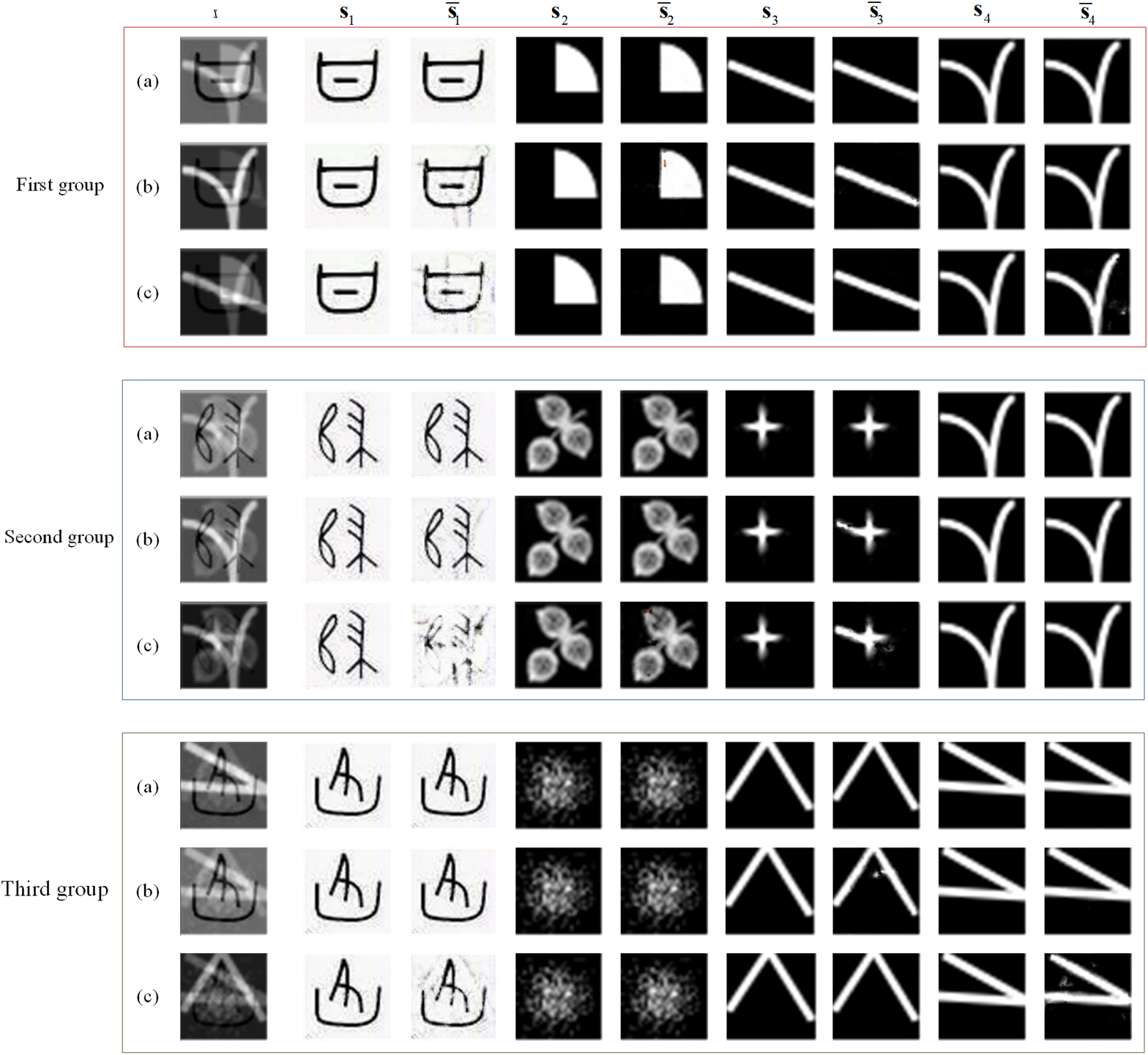}
	\caption{Testing results of PDualGAN approach for convolutive mixing ACC-occlusion images. Each group includes three different mixtures ((a), (b) and (c)) of four same images of different random weights. The $\mathbf{X}$ represents the mixture, $\mathbf{s}$$_1$, $\mathbf{s}$$_2$, $\mathbf{s}$$_3$, and $\mathbf{s}$$_4$ represent the four original ground-truth sources, $\mathbf{\bar s}$${_1}$, $\mathbf{\bar s}$${_2}$, $\mathbf{\bar s}$${_3}$, and $\mathbf{\bar s}$${_4}$ represent the corresponding estimated sources.}\label{Fig_7}
\end{figure*}
\begin{figure}[!h]\centering
	\includegraphics[width=7.5cm]{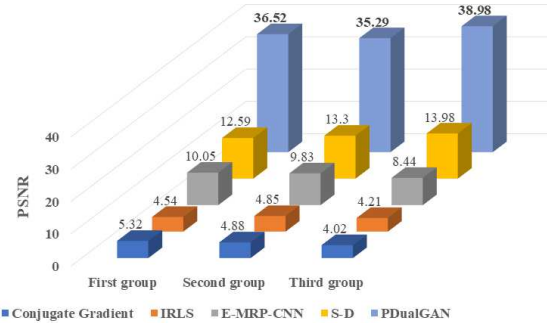}
	\caption{Average PSNR of the proposed PDualGAN as compared with four state-of-the-art convolutive model based algorithms.}\label{Fig_8}
\end{figure}

As observed from Fig. \ref{Fig_6}, the average PSNR of the proposed algorithm can reach 34.24 dB, outperforming the baseline algorithms EMD-ICA, EEMD-PCA-ICA, CEEDMAN-ICA, SSA-ICA and Gaussian WGAN of 8.64 dB, 8.42 dB, 8.22 dB, 9.25 dB, and 7.86 dB. TABLE \ref{tab2} demonstrates the average correlation that the PDualGAN algorithm can reach 0.9948 which outperforms the baseline algorithms EMD-ICA, EEMD-PCA-ICA, CEEDMAN-ICA, SSA-ICA and Gaussian WGAN of 0.2754, 0.3134, 0.2203, 0.3670 and 0.3754. Therefore, the PDualGAN algorithm achieves better results than the baseline instantaneous mixing model based algorithms in terms of both PSNR and correlation.

\subsection{PDualGAN for ACC-occlusion images.}

In the second set of simulations, we evaluate the separating performance of the proposed PDualGAN algorithm described in Algorithm 1 for convolutive mixtures on the ACC-occlusion datasets.

The testing is similar to section \emph{A}. we randomly select three groups of the convolutive mixtures and corresponding sources from the ACC-occlusion image datasets, each group shows the separating results from three different mixtures ((a), (b), (c) in Fig. \ref{Fig_7}), and each mixture is convolutive generated by random mixing matrixes and four same images with different random weights. Fig. \ref{Fig_7} shows the test results of separating four images from convolutive mixtures. We compared the PDualGAN algorithm with Conjugate Gradient, IRLS, S-D and E-MRP-CNN convolutive model based algorithms according to the different mixtures of Fig. \ref{Fig_7}. The comparison results of average PSNR values and average correlation obtained by these state-of-the-art algorithms are shown in Fig. \ref{Fig_8} and TABLE \ref{tab3} respectively.
\begin{table*}[htbp]
    \renewcommand{\arraystretch}{1.5}
    \renewcommand\tabcolsep{7pt}
 \caption{Average correlation of the proposed PDualGAN as compared with four state-of-the-art convolutive mixing algorithms.}
 \centering
 \label{tab3}
  \begin{tabular}{ccccccccccccccccccc}
   \hline\hline \\[-3mm]
             \multirow{2}{*}{Algorithm}
                                        & \multicolumn{3}{c}{First group} & \multicolumn{3}{c}{Second group} & \multicolumn{3}{c}{Third group}\\
                                        & (a) & (b) & (c) & (a) & (b) & (c) & (a) & (b) & (c)  \\ \hline
   \multicolumn{1}{l}{Conjugate Gradient} & 0.2395 & 0.2481  &0.2355  &  0.1127  & 0.1628  & 0.1512 & 0.2738 & 0.2522 & 0.2152        \\ \hline
   \multicolumn{1}{l}{IRLS}& 0.1513 & 0.1722  & 0.1371  &  0.0921  & 0.1144  & 0.1525 & 0.2314 & 0.2594 & 0.2182 \\ \hline
   \multicolumn{1}{l}{S-D}& 0.7422 & 0.6088  & 0.5977  &  0.7004  & 0.7936  & 0.5142 & 0.5754 & 0.7882 & 0.7621 \\ \hline
   \multicolumn{1}{l}{E-MRP-CNN}& 0.5188 & 0.2935 & 0.4715 & 0.4132 & 0.4485  & 0.3718 & 0.5711 & 0.3681 & 0.3575 \\ \hline
   \multicolumn{1}{l}{PDualGAN}& 0.9998 & 0.9918 & 0.9673 &  0.9806 & 0.9998  & 0.8492 & 0.9998 & 0.9958 & 0.9842
   \\[0.5ex]
   \hline\hline
  \end{tabular}
\end{table*}

Obviously, the average PSNR of the proposed algorithm can reach 38.98 dB, outperforming the baseline algorithms Conjugate Gradient, IRLS, S-D, and E-MRP-CNN of 5.32 dB, 4.85 dB, 13.98 dB, and 10.05 dB. TABLE \ref{tab3} demonstrates the average correlation that the PDualGAN algorithm can reach 0.9998 which outperforms the baseline algorithms Conjugate Gradient, IRLS, S-D, and E-MRP-CNN of 0.2738, 0.2594, 0.7621, and 0.5711.

\subsection{PDualGAN for speech signals.}

\begin{figure*}[htbp]\centering
	\includegraphics[width=11.5cm]{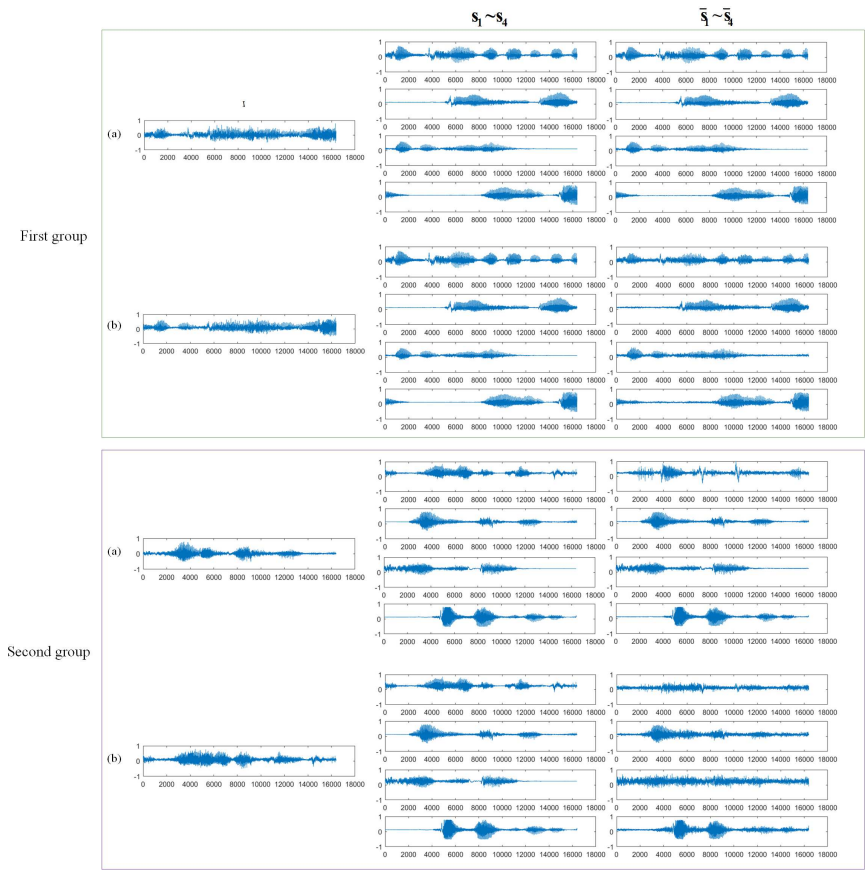}
	\caption{Testing results of PDualGAN approach for convolutive mixing speech signals. Each group including two different mixtures((a), (b)) of four same signals of different random weights. The first column is the mixtures, the second column is the corresponding original sources $\mathbf{s}$$_1$, $\mathbf{s}$$_2$, $\mathbf{s}$$_3$, and $\mathbf{s}$$_4$, and the third column is the corresponding estimated sources $\mathbf{\bar s}$${_1}$, $\mathbf{\bar s}$${_2}$, $\mathbf{\bar s}$${_3}$, and $\mathbf{\bar s}$${_4}$.}\label{Fig_9}
\end{figure*}
\begin{table*}[htbp]
    \renewcommand{\arraystretch}{1.7}
    \renewcommand\tabcolsep{9pt}
 \caption{Average correlation of the proposed PDualGAN as compared with four state-of-the-art convolutive mixing algorithms.}
 \centering
 \label{tab4}
  \begin{tabular}{ccccc}
   \hline\hline \\[-4mm]
             \multirow{2}{*}{Algorithm} & \multicolumn{2}{c}{First group} & \multicolumn{2}{c}{Second group} \\
                                        & (a) $\mathbf{s}$$_i$ and $\mathbf{\bar s}$$_i$ & (b) $\mathbf{s}$$_i$ and $\mathbf{\bar s}$$_i$ & (a) $\mathbf{s}$$_i$ and $\mathbf{\bar s}$$_i$  & (b) $\mathbf{s}$$_i$ and $\mathbf{\bar s}$$_i$\\ \hline

             \multicolumn{1}{l}{Conjugate Gradient}  &  0.3662 & 0.0733        &  0.1987 & 0.1681        \\ \hline
             \multicolumn{1}{l}{IRLS}  & 0.1588    & 0.2045              & 0.2943  & 0.1479             \\ \hline
             \multicolumn{1}{l}{S-D}  & 0.3548    & 0.5012              & 0.4988  & 0.4492             \\ \hline
             \multicolumn{1}{l}{E-MRP-CNN}  & 0.6735    & 0.5466              & 0.6022  & 0.6716   \\ \hline
             \multicolumn{1}{l}{PdualGAN}  & 0.9432    & 0.9656              & 0.7122  & 0.7041           \\[0.5ex]
   \hline\hline
  \end{tabular}
\end{table*}

In the third set of simulations, we evaluate the separating performance of the proposed PDualGAN algorithm described in Algorithm 1 for convolutive mixtures on the speech signal datasets.

The testing is similar to section \emph{A}. We randomly select two groups of the convolutive mixtures and corresponding sources from the Speech datasets, each group shows the separating results from two different mixtures ((a), (b) in Fig. \ref{Fig_9}), and each mixture is convolutive generated by four same signals with different random weights. Fig. \ref{Fig_9} shows the test results for convolutive mixing signals. Similarly, we compared the PDualGAN algorithm with Conjugate Gradient, IRLS, S-D and E-MRP-CNN algorithms according to Fig. \ref{Fig_9}. TABLE \ref{tab4} demonstrates the comparison results of the correlation obtained by the four state-of-the-art algorithms.

As shown in TABLE \ref{tab4}, the proposed PDualGAN algorithm can obtain an average correlation of up to 0.9656, which is better than the baseline algorithms Conjugate Gradient, IRLS, S-D, and E-MRP-CNN of 0.3662, 0.2943, 0.5012, and 0.6735.
\begin{table*}[htbp]
    \renewcommand{\arraystretch}{1.7}
    \renewcommand\tabcolsep{9pt}
 \caption{Average correlation of the proposed PDualGAN as compared with five state-of-the-art convolutive mixing algorithms.}
 \centering
 \label{tab5}
  \begin{tabular}{ccccc}
   \hline\hline \\[-4mm]
             \multirow{2}{*}{Algorithm} & \multicolumn{2}{c}{First group} & \multicolumn{2}{c}{Second group} \\
                                        & (a) $\mathbf{s}$$_i$ and $\mathbf{\bar s}$$_i$ & (b) $\mathbf{s}$$_i$ and $\mathbf{\bar s}$$_i$ & (a) $\mathbf{s}$$_i$ and $\mathbf{\bar s}$$_i$  & (b) $\mathbf{s}$$_i$ and $\mathbf{\bar s}$$_i$\\ \hline
             \multicolumn{1}{l}{EMD-ICA}  & 0.0181    & 0.0124    & 0.0159  & 0.0125           \\ \hline
             \multicolumn{1}{l}{EEMD-PCA-ICA}  & 0.0259    & 0.0242              & 0.0356  & 0.0301       \\ \hline
             \multicolumn{1}{l}{CEEDMAN-ICA}  & 0.0948    & 0.0812              & 0.0488  & 0.0392     \\ \hline
             \multicolumn{1}{l}{SSA-ICA}  &  0.0130   &  0.0154             & 0.0155  &  0.0101           \\ \hline
             \multicolumn{1}{l}{Gaussian WGAN}  &  0.1780   &  0.1154           & 0.1259  &  0.1982     \\ \hline
             \multicolumn{1}{l}{PdualGAN}  & 0.7432    & 0.7018              & 0.9743  & 0.9741         \\[0.5ex]
   \hline\hline
  \end{tabular}
\end{table*}

\begin{figure*}[!t]\centering
	\includegraphics[width=12cm]{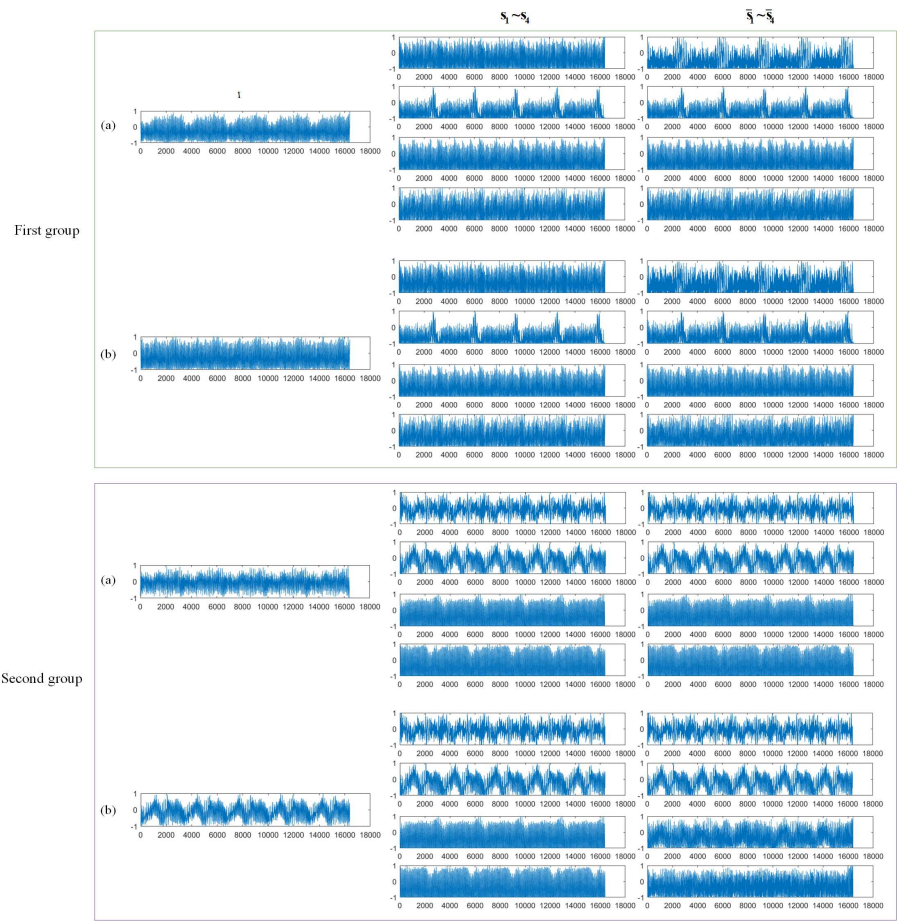}
	\caption{Testing results of PDualGAN approach for instantaneous mixing EEG signals. Each group including two different mixtures((a), (b)) of four same signals of different random weights. The first column is the mixtures, the second column is the corresponding original sources $\mathbf{s}$$_1$, $\mathbf{s}$$_2$, $\mathbf{s}$$_3$, and $\mathbf{s}$$_4$, and the third column is the corresponding estimated sources $\mathbf{\bar s}$${_1}$, $\mathbf{\bar s}$${_2}$, $\mathbf{\bar s}$${_3}$, and $\mathbf{\bar s}$${_4}$.}\label{Fig_10}
\end{figure*}
\subsection{PDualGAN for EEG signals.}
In the fourth set of simulations, we evaluate the separating performance of the proposed PDualGAN algorithm described in Algorithm 1 for instantaneous mixtures on the EEG signal datasets.

The testing is similar to section \emph{A}. We randomly select two groups of the instantaneous mixtures and corresponding sources from the EEG datasets, each group shows the separating results from two different mixtures ((a), (b) in Fig. \ref{Fig_10}), and each mixture is instantaneously mixing generated by four same signals with different random weights. Fig. \ref{Fig_10} shows the test results for instantaneous mixing signals. Similarly, we compared the PDualGAN algorithm with EMD-ICA, EEMD-PCA-ICA, CEEDMAN-ICA, SSA-ICA and Gaussian WGAN algorithms according to Fig. \ref{Fig_10}. TABLE \ref{tab5} demonstrates the comparison results of the correlation obtained by the five state-of-the-art algorithms.

Obviously, the average correlation that the PDualGAN algorithm can reach 0.9743 which outperforms the baseline algorithms EMD-ICA, EEMD-PCA-ICA, CEEDMAN-ICA, SSA-ICA, and Gaussian WGAN of 0.0181, 0.0356, 0.0948, 0.0155 and 0.1982.

The experimental results show that the proposed PDualGAN algorithm outperforms state-of-the-art algorithms in both PSNR and correlation on the four different datasets for the instantaneous mixtures and the convolutive mixtures, which shows the effectiveness of the proposed algorithm.
\section{Conclusion}
In this paper, a new algorithm for the problem of single channel blind source separation (SCBSS) has been presented. Our contributions are as follows:

\textbf{Model.} We have formulated a unified model for the instantaneous mixing model and convolutive mixing model for the research.

\textbf{Algorithm.} Based on the instantaneous mixing model and convolutive mixing model, we proposed a PDualGAN algorithm. \emph{N} DualGAN are trained simultaneously with mixtures and corresponding sources to realize one-to-multiple mapping, and Wasserstein generative adversarial networks gradient penalty(WGAN-GP) loss function is applied to improve the stability of the network.

The proposed algorithm can be used to both one-dimensional and two-dimensional signals, and different mixtures are applied to test the effectiveness and generalization performance.

\textbf{Datasets.}  We build the one-to-multiple datasets in the experiment which are composed of two parts: the mixtures and corresponding sources. Each mixture is generated by using randomly generated mixing matrixes and multiple sources with different weights. These datasets can be used for related research.

Numerical experiments show that the proposed PDualGAN algorithm performs well in separating multiple sources from instantaneous mixtures and convolutive mixtures which outperforms the EMD-ICA, EEMD-PCA-ICA, CEEDMAN-ICA, SSA-ICA, Conjugate Gradient, IRLS, S-D and DCSS algorithms.
\section*{Acknowledgment}
This study was supported by the China Scholarship Council under Grant [2020]1417, the Key Research and Development Project of Shanxi Province under Grant 201803D421035, the Natural Science Foundation for Young Scientists of Shanxi Province under Grant 201901D211313, the Research, Teaching, and Research Funding Project of Shanxi Province for Returned Overseas Students under Grant HGKY2019080. The authors would like to thank the anonymous reviewers for their constructive comments improving this article.

\ifCLASSOPTIONcaptionsoff
  \newpage
\fi

\end{document}